\documentclass[10pt,conference,a4paper]{IEEEtran}
\usepackage{cite}
\usepackage{amsmath,amssymb,amsfonts}
\usepackage{algorithmic}
\usepackage{graphicx}
\usepackage{textcomp}
\usepackage{xcolor}
\def\BibTeX{{\rm B\kern-.05em{\sc i\kern-.025em b}\kern-.08em
    T\kern-.1667em\lower.7ex\hbox{E}\kern-.125emX}}
    
\usepackage{url}        
\usepackage{booktabs}   
\usepackage{array}      
\usepackage{float}      
\begin{document}

\title{Pretraining Image Encoders without Reconstruction via Feature Prediction Loss}

\author{\IEEEauthorblockN{Gustav Grund Pihlgren}
\IEEEauthorblockA{\textit{EISLAB Machine Learning} \\
\textit{Lule\r{a} University of Technology}\\
Lule\r{a}, Sweden \\
gustav.pihlgren@ltu.se}
\and
\IEEEauthorblockN{Fredrik Sandin}
\IEEEauthorblockA{\textit{EISLAB Electronic Systems} \\
\textit{Lule\r{a} University of Technology}\\
Lule\r{a}, Sweden \\
fredrik.sandin@ltu.se}
\and
\IEEEauthorblockN{Marcus Liwicki}
\IEEEauthorblockA{\textit{EISLAB Machine Learning} \\
\textit{Lule\r{a} University of Technology}\\
Lule\r{a}, Sweden \\
marcus.liwicki@ltu.se}
}

\maketitle

\begin{abstract}


%
%
%
%
%
This work investigates three methods for calculating loss for autoencoder-based pretraining of image encoders:
The commonly used reconstruction loss, the more recently introduced deep perceptual similarity loss, and a feature prediction loss proposed here; the latter turning out to be the most efficient choice.
Standard auto-encoder pretraining for deep learning tasks is done by comparing the input image and the reconstructed image. 
Recent work shows that predictions based on embeddings generated by image autoencoders can be improved by training with perceptual loss, i.e., by adding a loss network after the decoding step.
So far the autoencoders trained with loss networks implemented an explicit comparison of the original and reconstructed images using the loss network.
However, given such a loss network we show that there is no need for the time-consuming task of decoding the entire image.
Instead, we propose to decode the features of the loss network, hence the name ``feature prediction loss''.
To evaluate this method we perform experiments on three standard publicly available datasets (LunarLander-v2, STL-10, and SVHN) and compare six different procedures for training image encoders (pixel-wise, perceptual similarity, and feature prediction losses; combined with two variations of image and feature encoding/decoding).
The embedding-based prediction results show that encoders trained with feature prediction loss is as good or better than those trained with the other two losses.
Additionally, the encoder is significantly faster to train using feature prediction loss in comparison to the other losses.
The method implementation used in this work is available online.\footnote{\url{https://github.com/guspih/Perceptual-Autoencoders}}



\begin{IEEEkeywords}
Autoencoder, Perceptual, Knowledge Distillation, Image Classification, Object Positioning, Embeddings
\end{IEEEkeywords}

\end{abstract} 

\section{Introduction}
\label{toc:introduction}




\begin{figure}[t!] 
    \centering
    \includegraphics[width=0.9\columnwidth]{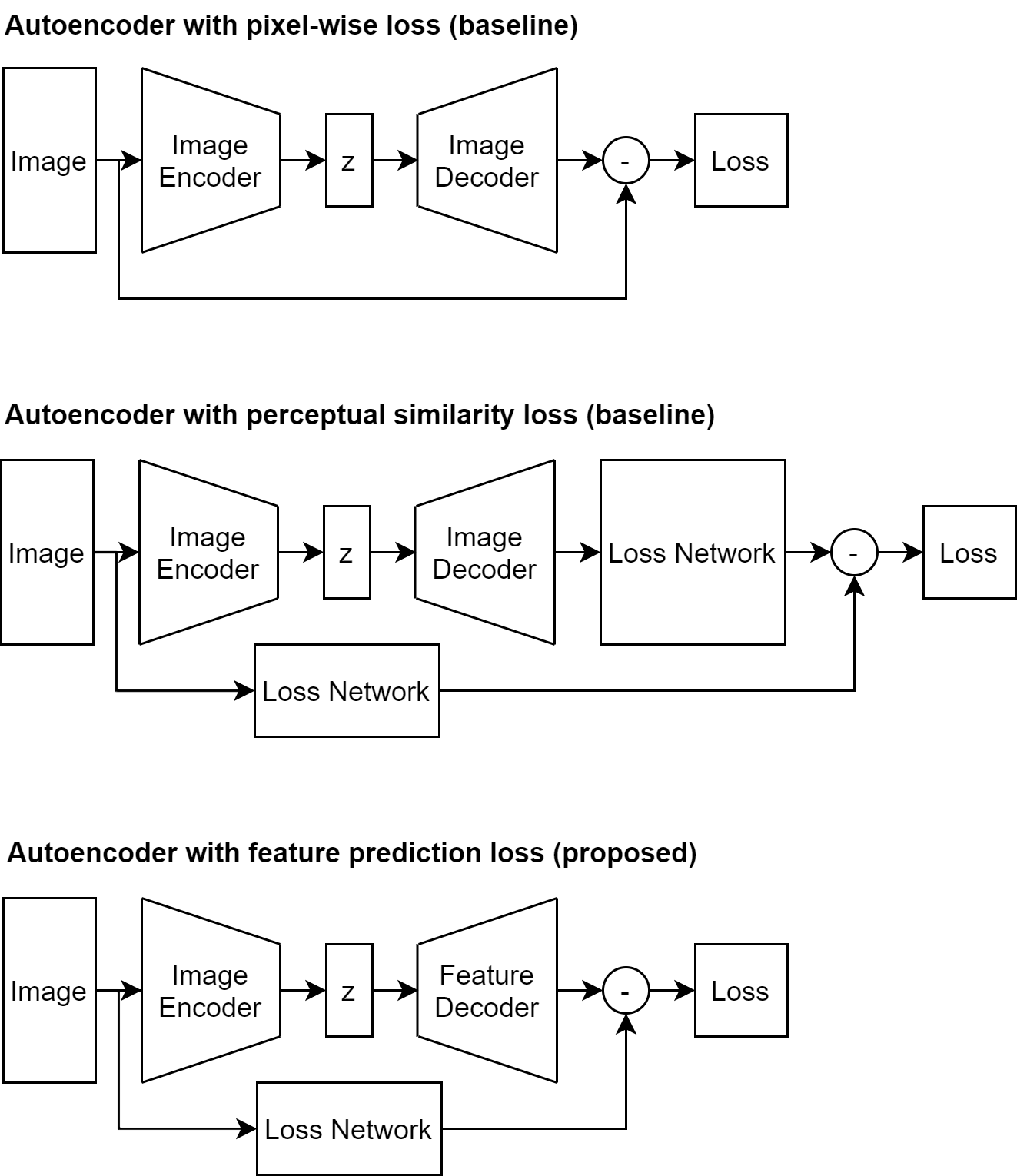}
    \caption{
    Three of the six evaluated image encoder pretraining procedures.
    }
    \label{fig:abstract_image}
\end{figure}

Deep perceptual loss is the use of the activations in a representation layer of a neural network (e.g., a classification network) to compute the loss of another machine learning model \cite{larsen2016autoencoding}. 
The network used to calculate the deep perceptual loss will be referred to here as the (perceptual) loss network.
The general principle of deep perceptual loss is to feed the output of the model to the loss network and use the activations of the loss network as the basis for the loss function being optimized.
This is a contrast to element-wise loss where the outputs are used directly as part of the loss function being optimized.

Since its inception, the use of deep perceptual loss has become more and more prominent in the image processing field.
This increased use is often motivated by their performance in comparison to element-wise calculated loss:
``Element-wise metrics are simple but not very suitable for image data, as they do not model the properties of human visual perception''~\cite{larsen2016autoencoding}. 
The idea that deep perceptual metrics better model human perception of image similarity was given further evidence in~\cite{zhang2018unreasonable}.
In the specific case of images represented by pixels, element-wise loss is called pixel-wise (PW) loss.

A prominent use of deep perceptual loss has been in the training of autoencoders.
Perceptually trained autoencoders have been used for, among other applications, image generation~\cite{larsen2016autoencoding}, style transfer and super-resolution~\cite{johnson2016style}, image segmentation~\cite{mosinska2017beyond}, and image classification and positioning~\cite{pihlgren2020improving}.

In all of the above works an autoencoder type network is used to reduce the input image into a
lower-dimensional 
embedding, from which the image can be reconstructed. 
The reconstruction loss 
is calculated, in whole or in part, from the differences of some features of the loss network 
given the original and reconstructed images as input. 
In short, the image is encoded and reconstructed, and then the reconstruction is compared to the original by how similar the features extracted by the loss network are.
This procedure is referred to as training with deep perceptual similarity (PS) loss.

This procedure is effective when the aim is 
to generate images since the decoder is trained to do that.
Autoencoders are also commonly used for feature learning and dimensionality reduction~\cite{goodfellow2016deeplearningbook}.
Reconstructing the image is not necessary for these tasks, but when using PW or PS loss it is needed anyways since both losses require the reconstruction. 
PS loss has been shown to perform better than PW loss on this task~\cite{pihlgren2020improving}.
However, when a loss network is available the decoding part of the training protocol can be avoided altogether.

This work proposes an alternative to training autoencoders with PS loss, dubbed feature prediction (FP) loss, which does not require reconstruction of the image.
The training procedure for FP loss is as follows.
First, a feature extraction of the original image is generated with the loss network.
Second, the image is embedded using an encoder.
Third, the feature extraction is predicted from the embedding with the decoder.
%
This method implies that the loss network is used as a teacher network for knowledge distillation~\cite{hinton2015distilling} rather than using the deep perceptual loss.
Though in this case, the knowledge being distilled is the deep features of the teacher network rather than the outputs.
For consistency, we refer to the neural network used for feature extraction and loss calculation as the loss network regardless of how it is used.

The proposed procedure is compared to training with both PW loss and PS loss.
This is done by generating embeddings with the trained autoencoders and evaluating how good the embeddings are.
The performance of the embeddings is measured by training Multi-Layer Perceptrons (MLP) for predicting the labels of the dataset from the embeddings.
Three different datasets are tested with labels for classification or object positioning.
The two baseline autoencoder procedures as well as the proposed procedure is shown in Fig.~\ref{fig:abstract_image}.

\subsection*{Contribution}

The major contribution of this work is proposing feature prediction (FP) loss as an alternative to deep perceptual similarity (PS) loss when 
pretraining image encoders.
Therefore, we
\begin{itemize}
    \item compare six different procedures for autoencoding, based on three ways of calculating loss; pixel-wise (PW), PS, and FP loss; combined with different variations of image and feature encoding/decoding.
    \item test the procedures 
    on three different datasets and with three different sizes of the latent space.
    \item show that encoders are significantly faster to train with FP loss than the other two losses.
    \item demonstrate that the embeddings created by encoders trained with FP loss are equal or better, for prediction on all three datasets, than using the other two losses.
\end{itemize}

\section{Related Work}
\label{toc:related_Work}

The autoencoder is a prominent neural network architecture that has been used in some form since the 1980s~\cite{rumelhart1985learning, ballard1987modular}.
Autoencoders are generally trained in an unsupervised fashion by making the target output the same or similar to the input and minimizing the difference between the two.
This is made non-trivial by having a so-called latent space in between the input and output, where the number of dimensions is much lower than that of the input and output.
The latent space thus constitutes a bottleneck, and all data from the input that is needed for reconstruction will have to be compressed into the latent space. 
The part of the network that takes the input and reduces it into the latent space is called the encoder, and the part that reconstructs the output from the latent space is called the decoder.

In addition to dimensionality reduction, autoencoders have been used for a host of different task including generative modelling~\cite{yuhuai2016generative}, denoising~\cite{vincent2008denoising}, generating sparse representations~\cite{ranzato2007sparse}, and anomaly detection~\cite{hawkins2002outlier}.

Deep perceptual loss, in the form of optimizing the input of a neural network with respect to the activations generated by that neural network, was first introduced in the field of explainable AI.
In that field, deep perceptual loss was used originally to visualize a network's perceived optimal input image for some class \cite{simonyan2014deep} or some individual units \cite{yosinski2015understanding}.

Simultaneously with the introduction of deep perceptual loss in explainable AI, it was also introduced as a method to generate adversarial examples~\cite{szegedy2014intriguing}.
Adversarial examples are inputs to machine learning models which are constructed to produce the incorrect model outputs even though they are almost indistinguishable from inputs resulting in correct correct outputs.

Another deep perceptual loss-based approach, Generative Adversarial Networks (GAN), was introduced soon after~\cite{goodfellow2014generative}.
In GANs a generator network is trained to generate images that fools a discriminator network that the image was taken from the ground truth.
In this case, the discriminator acts as a loss network for the generator.

Deep perceptual loss was first used with autoencoders when the GAN was combined with the Variational Autoencoder (VAE) to create the VAE-GAN~\cite{larsen2016autoencoding}.
In the VAE-GAN the decoder network of a VAE is the same as the generator network of a GAN.
It's a VAE that attempts to reconstruct images to fool the discriminator.
In addition to the deep perceptual loss for fooling the discriminator, there is an additional deep perceptual loss generated by feature extraction from the discriminator when its given both the original and reconstructed images.
This second loss is therefore equivalent to deep perceptual similarity loss as described in the Introduction.

In~\cite{dosovitskiy2016generating} a pretrained computer vision model replaced the discriminator as the loss network.
This removed the need for the time-consuming task of training the discriminator.

Another method that similarly uses a pretrained network to optimize another model is knowledge distillation~\cite{hinton2015distilling}.
This method most commonly uses the prediction values of a pretrained network either as a replacement for or together with the ground truth when training a new model.
In this setup, the pretrained model is referred to as the teacher and the model being trained is the student.
Knowledge distillation has been shown to give faster training time and higher performance of the student model than training using only the ground truth~\cite{yim2017gift}.

\section{Datasets}
\label{toc:datasets}

This work uses three different datasets for the evaluation of the methods:
The LunarLander-v2 collection, STL-10~\cite{coates2011analysis}, and SVHN~\cite{netzer2011reading}.
The datasets and how this work makes use of them are described below.

\subsection{LunarLander-v2 collection}
The LunarLander-v2 collection is a collection of images from the LunarLander-v2 environment of the OpenAI Gym~\cite{brockman2016openai}.
The images were collected from three runs of $700$ rollouts each.
All rollouts were made for $150$ timesteps with a random policy controlling the lander.
Each of the three runs therefore collected $105000$ images which were scaled down to $64\times64$ pixels as well as the positions of the lander.
For the second and third runs the images where the lander is outside the screen were removed.
This removed roughly $10\%$ of the images.

The first run is used for unsupervised training and validation of the autoencoders.
The second run is used for training and validation of object positioning of the lander.
The third run is used for testing of object positioning.

\subsection{STL-10}
The STL-10 dataset is $108500$ photos of animals and vehicles acquired from the larger ImageNet~\cite{deng2009imagenet} dataset scaled to $96\times96$ pixels.
$100000$ of the images are unlabeled and are, in this work, used for training and validation of the autoencoder.
$500$ of the images are labeled and intended for training, those are used for training and validation of classification.
$8000$ of the images are labeled and intended for testing, and are used for testing of classification.
The labeled images are divided into $10$ classes of animals and vehicles.
The unlabeled data contains images of animals and vehicles both inside and outside of the $10$ classes.

This dataset is being used for comparison of the models and as such this work does not follow the test protocol provided by the creators of the dataset.
Results achieved in this work can therefore not be directly compared to results that are achieved when following that protocol.

\subsection{SVHN}
The SVHN dataset is photos of house numbers with $630420$ individual digits that have been labeled and given bounding boxes.
The dataset is available with either the original photos or with the individual digits cropped out and scaled to $32\times32$ pixels.
This work uses the cropped and scaled digits.
In the dataset $73257$ of the digit are marked for training, $26032$ for testing, and $531131$ as extra.
The extra images are used for training and validation of autoencoders in this work.
The training images are used for training and validation of classification.
The testing images are used for testing of classification.

\section{Loss Calculation}
\label{toc:perceptual_loss}

In this work loss is calculated in three different ways: 
PW, PS, and FP loss which are described in Eq.~\ref{eq:pixel_wise}, Eq.~\ref{eq:perceptual_similarity}, and Eq.~\ref{eq:feature_prediction} respectively.
In the equations $X$ is an image with $n$ pixels being embedded into a latent space with $m$ dimensions, $en$ is an encoder, $de$ is a decoder, $p$ is a loss network, and $f$ is a loss function (like square error or cross-entropy).

\begin{equation}
    E = \sum^n_{k=1} f(X_k, de(en(X))_k)
    \label{eq:pixel_wise}
\end{equation}

\begin{equation}
    E = \sum^m_{k=1} f(p(X)_k, p(de(en(X)))_k)
    \label{eq:perceptual_similarity}
\end{equation}

\begin{equation}
    E = \sum^m_{k=1} f(p(X), de(en(X))_k)
    \label{eq:feature_prediction}
\end{equation}

In Eq.~\ref{eq:pixel_wise} and Eq.~\ref{eq:perceptual_similarity} the decoder outputs an image of size $n$.
In Eq.~\ref{eq:feature_prediction} the decoder outputs a vector of the same size as the feature extraction from $p$.

\begin{figure}[ht] 
    \centering
    \includegraphics[width=\columnwidth]{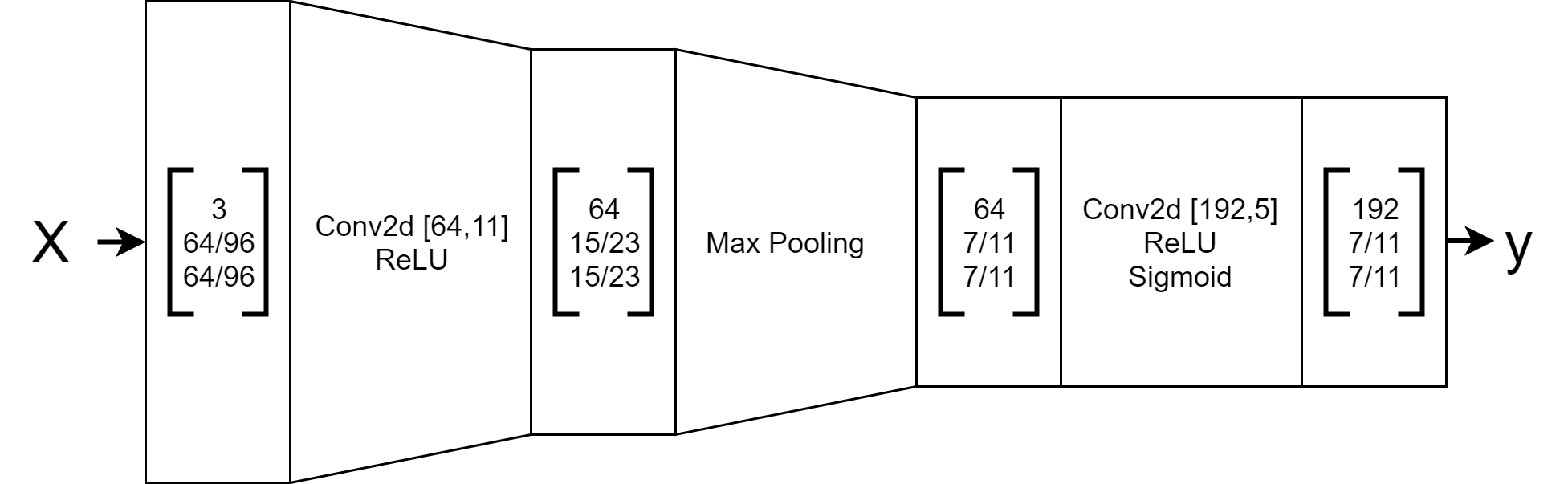}
    \caption{The parts of a pretrained AlexNet that were used for calculating and backpropagating the deep perceptual loss~\cite{pihlgren2020improving}.}
    \label{fig:image_alexnet_loss}
\end{figure}

The loss network ($p$) used in this work is AlexNet~\cite{krizhevsky2014one} pretrained on ImageNet~\cite{deng2009imagenet}.
The feature extraction ($y$) from $p$ is the activations after the second ReLU layer.
The values of $y$ are normalized between $0$ and $1$ using the sigmoid function.
This is the same setup as in~\cite{pihlgren2020improving}.
The loss network is visualized in Fig.~\ref{fig:image_alexnet_loss}.

\section{Architecture}
\label{toc:autoencoder_architecture}

There are two different encoders and two different decoders used in this work to encode and decode either images or features.

\begin{figure}[ht] 
    \centering
    \includegraphics[width=0.75\columnwidth]{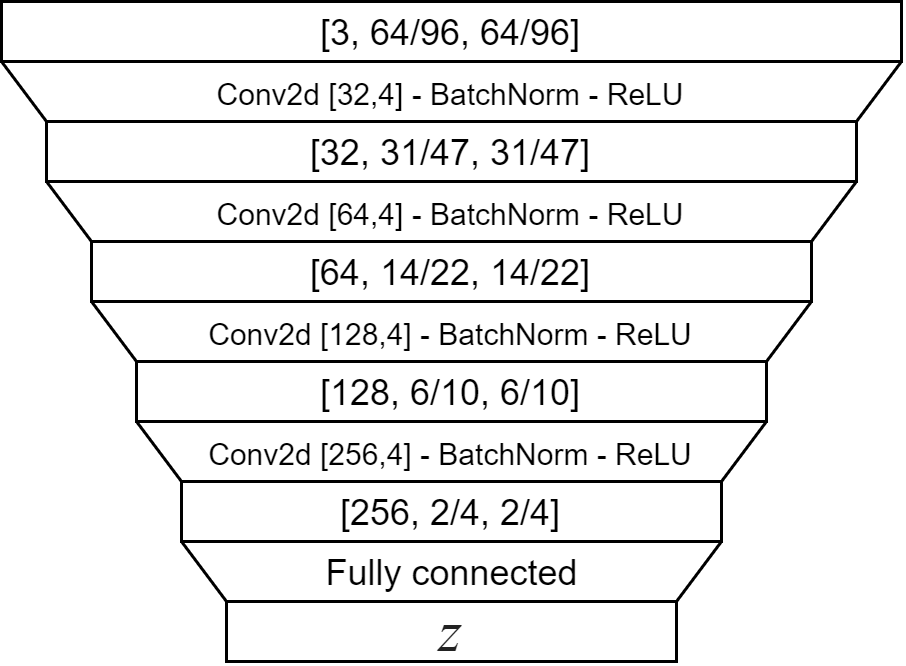}
    \caption{The convolutional image encoder used in this work.}
    \label{fig:image_encoder}
\end{figure}

\begin{figure}[ht] 
    \centering
    \includegraphics[width=0.75\columnwidth]{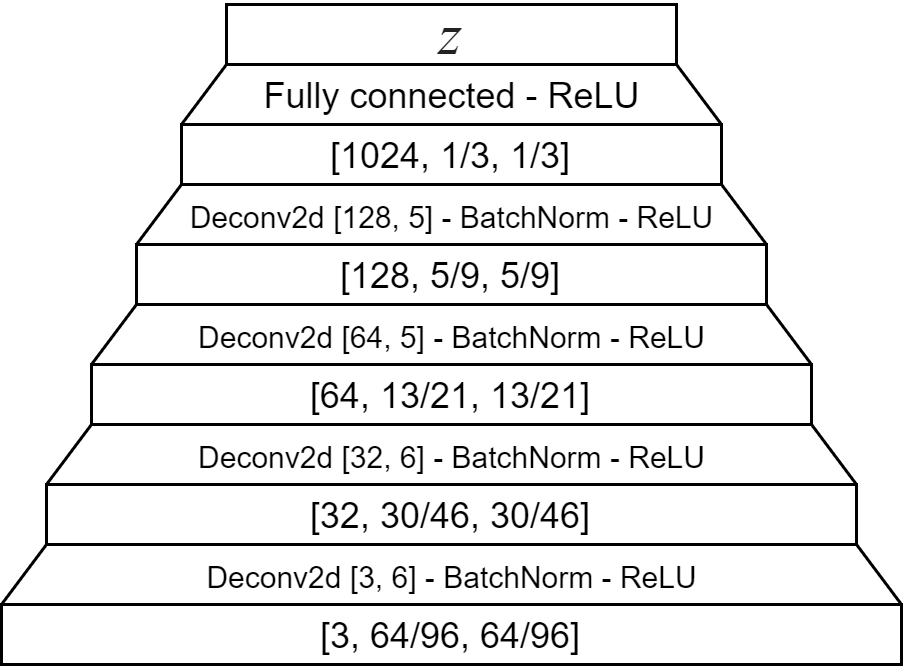}
    \caption{The convolutional image decoder used in this work.}
    \label{fig:image_decoder}
\end{figure}

The image encoder and image decoder are convolutional and can be seen in detail in Fig.~\ref{fig:image_encoder} and Fig~\ref{fig:image_decoder}.
In both figures, the boxes contain the intermediate layer sizes in the shape $[channels, width, height]$ with $/$ separating different sizes depending on the size of the input image.
In between the boxes are the functions that are applied to the data.
For the convolutional and deconvolutional layers the parameters are presented in shape $[nr\text{ }of\text{ }kernels, kernel\text{ }size]$.
The stride of all convolutional and deconvolutional layers is 2.

The feature encoder and feature decoder are Multi-Layer Perceptrons (MLP) with a single hidden layer with size 2048.
For the encoder, the input size is the size of the extracted features and the output size is the size of the latent space.
It's the other way around for the decoder.

Four different autoencoder architectures are used in this work given by all combinations of the encoders and decoders above.
The autoencoders with an image encoder takes the plain images as input.
The autoencoders with a feature encoder takes the feature extraction ($y$) from AlexNet of the plain image as input.
The autoencoders with an image decoder are trained either with PW loss (Eq.~\ref{eq:pixel_wise}) or PS loss (Eq.~\ref{eq:perceptual_similarity}).
The autoencoders with a feature decoder are trained with FP loss (Eq.~\ref{eq:feature_prediction}).

\section{Experiment Procedure}
\label{toc:experiment_procedure}

This work compares six different procedures; two different architectures for each of the three losses.
They are referred to according to the following scheme ``(Encoder type)-(Decoder type)-(Loss function)''.
Where ``I'' represents image encoder or decoder, ``F'' feature encoder or decoder, ``PW'' pixel-wise loss, ``PS'' deep perceptual similarity loss, and ``FP'' feature prediction loss.
The six procedures are listed below.
\begin{itemize}
    \item I-I-PW: Normal image autoencoder trained with PW loss (baseline).
    \item I-I-PS: Normal image autoencoder trained with PS loss (baseline).
    \item F-I-PW: Autoencoder that encodes features of the loss network and decodes the image trained with PW loss. Can be viewed as using the loss network for transfer learning.
    \item F-I-PS: Autoencoder that encodes features of the loss network and decodes the image trained with PS loss. Can be viewed as using the loss network for transfer learning.
    \item I-F-FP: Autoencoder that encodes the image and decodes the features of the loss network trained with FP loss. The proposed alternative to training with PS loss.
    \item F-F-FP: Autoencoder trained with FP loss to simply encode and decode the features of the loss network.
\end{itemize}

For each dataset, each procedure is run three times with different values of $z$: $64$, $128$, and $256$.
The autoencoders are trained for $50$ epochs and the final model is from the epoch with the lowest validation loss.
Of the data used for autoencoder training and validation, $20\%$ are used for validation.

\begin{figure}[ht] 
    \centering
    \includegraphics[width=0.9\columnwidth]{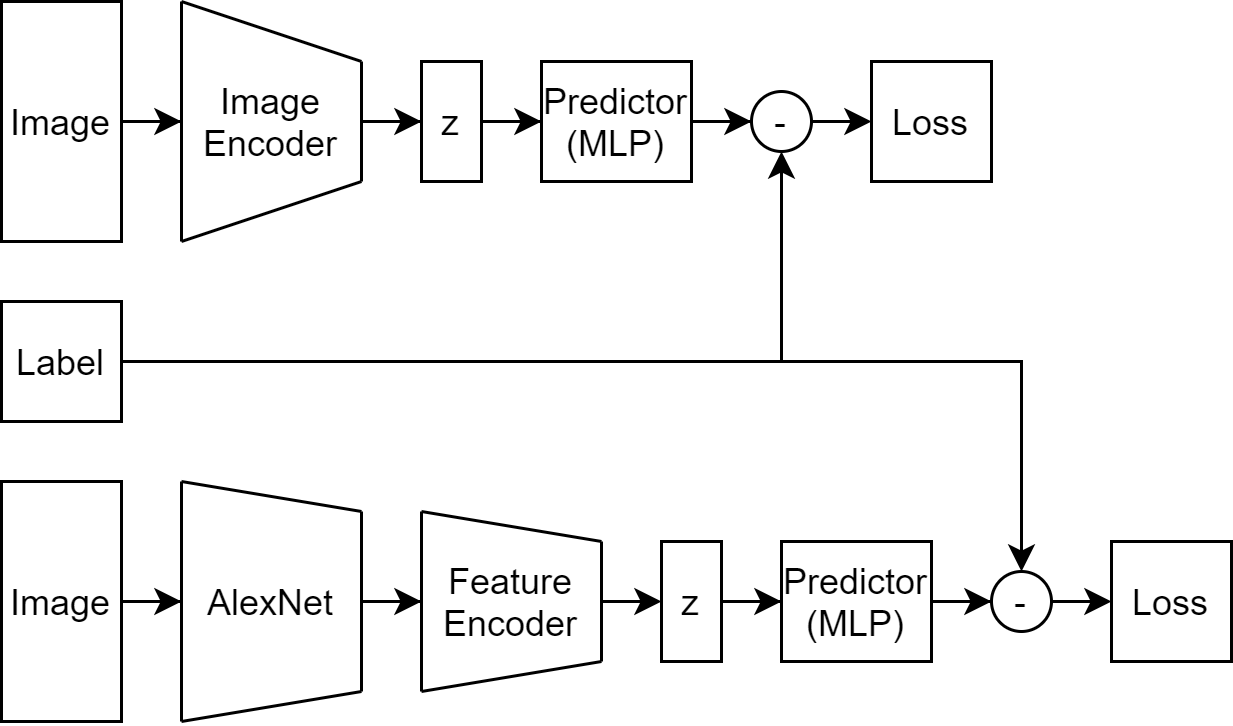}
    \caption{The convolutional image decoder used in this work.}
    \label{fig:training_setup}
\end{figure}

For each trained autoencoder $7$ predictor MLPs with different architectures were trained to do classification or object positioning.
The MLPs had input size $z$ and output size $10$ or $2$ for classification or object positioning, respectively.
The MLPs varied in the number and size of hidden layers:
No hidden layer, one layer with size $32$ or $64$, or two hidden layers with sizes $[32,32]$, $[64,32]$, $[64,64]$, or $[128,128]$. 
The MLPs were trained for $500$ epochs and the final model is from the epoch with the lowest validation loss.
$20\%$ of the classification and object positioning data are used for validation.
The training setup for the training of the predictor MLPs is shown in Fig~\ref{fig:training_setup}.

For each autoencoder, the MLP with the lowest validation loss was tested on the test set.
For classification, the test measure used is accuracy, while for object positioning it is the distance between the predicted position and the actual position.
The complete experiment procedure was repeated $4$ times.

\section{Results}
\label{toc:results}

\begin{figure}[ht!] 
    \centering
    \includegraphics[width=\columnwidth]{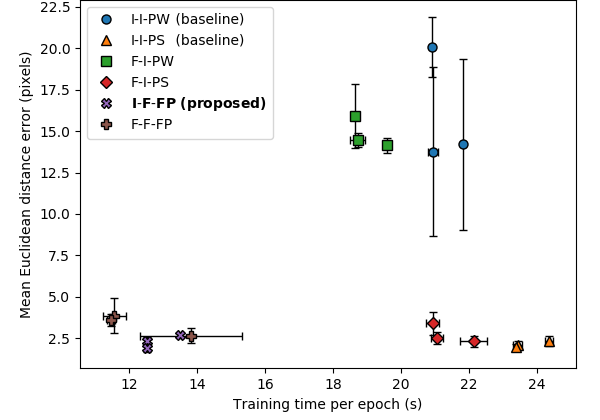}
    \caption{The mean Euclidean distance between the predicted position and actual position on the LunarLander-v2 collection test set versus the training time for the six different procedures for three different values of $z$.}
    \label{fig:lunarlander_results}
\end{figure}

\begin{figure}[ht!] 
    \centering
    \includegraphics[width=\columnwidth]{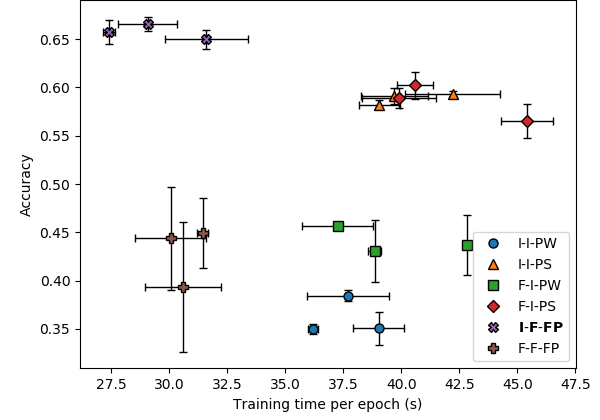}
    \caption{The accuracy on the STL-10 test set versus the training time for the six different procedures for three different values of $z$.}
    \label{fig:stl10_results}
\end{figure}

\begin{figure}[ht!] 
    \centering
    \includegraphics[width=\columnwidth]{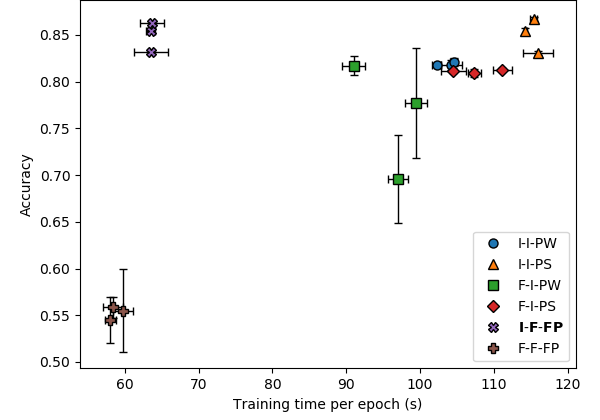}
    \caption{The accuracy on the SVHN test set versus the training time for the six different procedures for three different values of $z$.}
    \label{fig:svhn_results}
\end{figure}

The results consist of the performance and training time of the procedures.
The performance of each procedure is measured by the performances on the test sets for the autoencoder's MLP with the lowest validation loss.
The training time of the procedures are measured in the training time per epoch for each of the different autoencoders.
Fig.~\ref{fig:lunarlander_results},~\ref{fig:stl10_results}, and~\ref{fig:svhn_results} show these results for the LunarLander-v2 collection, STL-10, and SVHN datasets, respectively.

Each procedure has a color and shape, and has one point for each of the three $z$ values tested.
The data shown are the mean and uncorrected sample standard deviation for both performance and training time.

\begin{figure}[htb!] 
    \centering
    \includegraphics[width=\columnwidth]{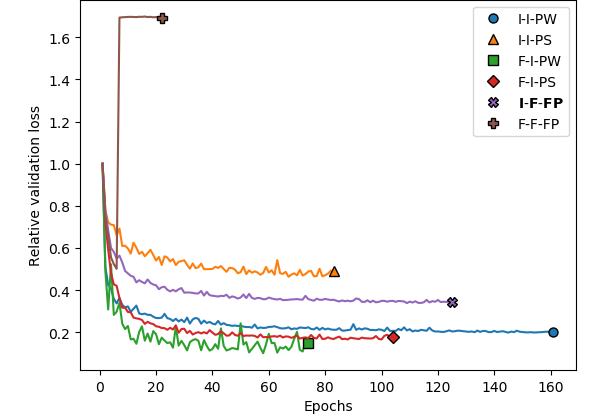}
    \caption{Convergence of autoencoder training of the six procedures on SVHN with $z=64$.}
    \label{fig:convergence}
\end{figure}

When training a model it is not only the training time per epoch that is important but also the number of epochs until the training converges.
Even when taking the convergence rates into account the difference in training time between procedures mimic the differences in training time per epoch.
However, when convergence rate is taken into account the variances in training time increases hugely.
To begin with F-F-FP has a convergence rate similar to those of the other models, but around epoch $15$ it collapses into a state with high validation loss which it cannot recover from.
This collapse combined with our use of early stopping gives F-F-FP a total training time of about $50\%$ of I-F-FP. Convergence can be seen in Fig.~\ref{fig:convergence} where autoencoders were trained using all six procedures on SVHN with $z=64$.
Since different losses are used for the different procedures the reported validation loss in the figure has been scaled such that each training procedure starts with a loss of $1$ after the first epoch.
For the convergence trials, training ran until no better validation loss could be found for $15$ epochs.

The embeddings can be calculated once before training the classification or object positioning MLPs.
This means that the effect of the choice of procedure on the training time of classification and object positioning is negligible.

During inference the only difference between the procedures is which of the two encoders that are used; image encoder or feature extraction followed by feature encoding.
The difference in encoding time between the two was negligible, especially when the additional time for prediction based on those encodings are taken into account.

For completeness, an additional test on SVHN was performed for each procedure to closer resemble the settings of~\cite{netzer2011reading}, where the dataset was introduced.
This test uses a higher dimensionality latent-space as well as a lot more data to train the MLPs.
In the original tests a more difficult setup was used where the data used for unsupervised pretraining of the autoencoders and supervised training of the MLPs were disjoint to mimic potential real use cases better.
In this trial, the $z$ value was set to $500$ and the extra data rather than the data marked for training was used to train the MLPs.
While this still doesn't quite match the setup of the original work as we use another autoencoder architecture, no hyperparameter search, and MLPs instead of linear SVM, the more similar settings give similar results.
While the intent of this work is to show that training autoencoders for embedding using PS loss can be perfromed faster using FP loss, it is still reassuring to see that the method performs comparably with the original work.
These results compared to the results in~\cite{netzer2011reading} can be found in Table~\ref{tab:svhn}.

Due to time constraints, this trial was only performed once.

\begin{table}[htb]
    \caption[]{
    Performance on SVHN with $z=500$ and the extra data used for MLP training compared to the baseline results.
    }
    \label{tab:svhn}
    \begin{center}
    \begin{tabular}{l l l l}
    \toprule
        Model                                   & Accuracy \\\hline
        I-I-PW (baseline)                       & $0.891$ \\
        I-I-PS (baseline)                       & $0.913$ \\
        I-F-FP (proposed)                       & $0.892$ \\
        Original work~\cite{netzer2011reading}  & $0.897$ \\
    \bottomrule
    \end{tabular}
    \end{center}
\end{table}



\section{Analysis}
\label{toc:analysis}

On the LunarLander-v2 collection, the PW loss-based procedures perform significantly worse than the other procedures.
All the remaining procedures perform comparably with a significant gap in training time.
The FP loss-based procedures have only around $60\%$ of the training time of the PS loss-based procedures.
The gap in performance between PW and the other losses is likely due to the task's dependence on the lander, a small but visually distinct feature of the image.
Its small size gives it a small contribution to PW loss while its visual distinction gives it a significant contribution to the features which the other losses depend on.

On STL-10 there are three distinct levels of performance.
The I-F-FP procedure has the best performance with just above $65\%$ accuracy.
The two PS loss-based methods come in second with just below $60\%$ accuracy.
The three remaining procedures all fall in the span between $35\%$ and $50\%$ accuracy.
Most of the procedures that use the loss network perform well on STL-10, which is not surprising since the dataset is a subset of ImageNet which is the dataset that the loss network has been trained on.
It is notable though that the F-F-FP procedure performs poorly despite its purpose being essentially autoencoding the features from the loss network.
With regards to training time, the FP loss-based procedures are faster with only about $75\%$ of the training time of the other procedures.
The lesser time gained by using FP loss on STL-10 in comparison to the other datasets could be due to the increased size of the images.

On SVHN all methods except F-F-FP performed similarly well, with I-F-FP and I-I-PS peaking at slightly over $85\%$ accuracy.
F-F-FP achieves around $55\%$ accuracy on average and the remaining procedures achieve around $80\%$.
It should be noted that for two values of $z$ F-I-PW performs worse by $5$ and $10$ percentage points respectively.
Again on this dataset it is obvious that the FP based procedures have significantly shorter training time, with I-F-FP taking only $55\%$ of the time of I-I-PS while having almost the exact same performance.

All methods except F-F-FP seem to converge similarly fast and reach a close to optimal validation loss within $50$ epochs.
However, as has been pointed out in~\cite{alberti2017pitfall}, better reconstruction loss for an autoencoder does not necessarily imply embeddings that give better performance.
While this might also apply for deep perceptual similarity loss or for feature prediction loss, further research in this direction is needed.


\section{Discussion and Conclusion}
\label{toc:discussion}

The most notable result is that the I-F-FP procedure is both among the two fastest and the two best performing for all datasets.
On SVHN and STL-10 it is the only method that is both among the fastest and the best performing.
This shows that the proposed FP loss is faster for pretraining image encoders than PS loss and sometimes also outperforms it.

However, the other FP loss based procedure, F-F-FP performs significantly worse and is in the bottom three for STL-10 and by far the worst on SVHN.
This is interesting since the only difference between the two procedures is that F-F-FP uses transfer learning by encoding features taken from AlexNet rather than the original image.
This would imply that encoding the original image such that the AlexNet features can be decoded gives embeddings with more task-relevant information than autoencoding the features directly.
Furthermore, this issue is not present in the other transfer learning procedures F-I-PW and F-I-PS, which performs comparably to their non-transfer learning counterparts I-I-PW and I-I-PS on all datasets.

Another potential reason for the increased performance could be that the feature decoder resembles the predictor MLPs more closely and thus the information is encoded in a way that is easier for a shallow MLP to extract.
However the poor performance of F-F-FP is a strong argument against this case.
Likely the use of the loss network is the significant factor for performance as I-F-FP and I-I-PS have similar performance on all datasets.

As can be seen in Table~\ref{tab:sota} the proposed method is equal to or better than the two baseline procedures for autoencoder training.
However, no procedure comes close to the state-of-the-art results for the datasets where those are available.
Furthermore, on SVHN a simple CNN, consisting of the encoder in Fig.~\ref{fig:image_encoder} followed by an MLP with a single hidden layer of 256 neurons, outperforms all procedures.
This should not be interpreted as evidence that the procedures evaluated in this work are poor.
The purpose of this work is to evaluate different methods for pretraining image encoders in terms of how useful the generated embeddings are for prediction.
This work, therefore, uses the datasets, not to achieve state-of-the-art results on those datasets, but to evaluate how good the embeddings generated by the different procedures are.

\begin{table}[htb]
    \caption[]{
    Performance of the compared procedures, including also a simple CNN and the state-of-the-art for reference
    }
    \label{tab:sota}
    \begin{center}
    \begin{tabular}{l l l l}
    \toprule
        Model               & LunarLander       & STL-10        & SVHN \\\hline
        I-I-PW (baseline)   & $13.76\pm5.08$    & $0.38\pm0.00$ & $0.82\pm0.00$     \\
        I-I-PS (baseline)   & $1.96\pm0.18$     & $0.59\pm0.00$ & $0.87\pm0.00$     \\
        I-F-FP (proposed)   & $1.92\pm0.22$     & $0.67\pm0.01$ & $0.86\pm0.00$     \\
        Simple CNN          & $8.10\pm0.01$     & $0.48\pm0.03$ & $0.89\pm0.01$     \\
        SOTA & --- & $0.94$~\cite{berthelot2019mixmatch} & $0.99$~\cite{cubuk2018autoaugment} \\
    \bottomrule
    \end{tabular}
    \end{center}
\end{table}

While FP would seem to be the better alternative to PS, this is only the case when the encoder is going to be used as a pretrained part of a prediction system.
However, one of the major uses of PS loss is image generation.
FP loss is ill-suited for this task as an image generator would have to be trained in addition to the autoencoder, and there is no guarantee that the embeddings are well suited for image generation as this is not 
part of the loss.
With PS the image generator can be trained as part of the procedure by letting the decoder fill this role.
However, FP is not meant to replace PS but rather it is an alternative encoder pretraining procedure when image generation is not the purpose of the system.

While this work shows that FP can be a good alternative to PS many factors remains to be investigated, such as:
The optimal architectures for the procedures; which loss network to use and where to make the feature extraction; training with multiple losses simultaneously; testing on further datasets; testing the embeddings of FP for image reconstruction and generation; investigating domains other than computer vision, and more.


\bibliographystyle{IEEEtran}
\bibliography{biblio}

\end{document}